# Coincidences and the Encounter Problem: A Formal Account

**Jean-Louis Dessalles (dessalles@enst.fr)**
TELECOM ParisTech, 46 rue Barrault
F-75013 Paris, France

**Abstract**

Individuals have an intuitive perception of what makes a good coincidence. Though the sensitivity to coincidences has often been presented as resulting from an erroneous assessment of probability, it appears to be a genuine competence, based on non-trivial computations. The model presented here suggests that coincidences occur when subjects perceive *complexity drops*. Co-occurring events are, together, simpler than if considered separately. This model leads to a possible redefinition of subjective probability.

**Keywords:** coincidence; complexity; probability; unexpectedness; surprise.

## Coincidences

Suicides are banal events: About thirty people commit suicide every day in a country like France. Yet, a French national newspaper, *Le Figaro*, reported two anonymous suicides in its March 20, 2004 edition, insisting on their similarity: Two late middle aged women who did not know each other walked in elegant dress into the sea and drowned, almost simultaneously early in the morning, just a few kilometers away from one another. The coincidence was so strange that the news was subsequently analyzed on a national radio.

The famous parallel between Abraham Lincoln and John F. Kennedy's fates leaves no one indifferent, even if one knows that the coincidence should not be regarded as unlikely (Kern & Brow, 2001). Their election to Congress and then as president of the United States, the births of their successors, the births of their assassins occurred on the same years, one century apart. Both successors were named Johnson. Both assassins were themselves assassinated before being tried.

Two Parisian colleagues running into each other in a small village close to Coban, Guatemala, perceive the coincidence as incredible and urge to tell the event to their friends.

Studying why and how human beings are fascinated by coincidences constitutes an important chapter of cognitive science. Coincidence avoidance has definite implications for decision making, especially in science and in court of law. Moreover, people can accurately assess the strength of coincidences (Griffiths & Tenenbaum, 2001; 2007), which means that they are able to capture complex relations between events.

The human sensitivity to coincidences is an embarrassment for most current models of cognition. Some authors consider it as mere marginal malfunction of an otherwise functional general ability to assess the likelihood of events. The malfunction would be due to representativeness bias (Kahneman & Tversky, 1972), to erroneous prior knowledge (Bar-Hillel, 1980) or to failure to consider proper alternatives (Diaconis & Mosteller, 1989; Falk, 1989; Tversky & Kohler, 1994). Other authors see in the perception of coincidences a fundamental device for concept learning and theory formation (Griffiths & Tenenbaum, 2007). The aim of the present paper is not to settle the issue, although our model clearly depicts the sensitivity to coincidences as a genuine competence rather than as the result of a malfunction. Our contribution, which builds on recent studies on cognitive complexity, is limited to showing that the perception of coincidences obeys definite, formal laws. This mere fact, if correct, may have some implications for our understanding of human cognition, especially by shedding new light on the notion of subjective probability.

Though coincidences are systematically experienced as improbable by subjects, their relation to probability is notoriously unclear: Among events of same probability, some may appear coincidental and others not (Griffiths & Tenenbaum, 2007). For instance, children's attention is grabbed when the family car reaches 66666 km on the clock, but they do not care when they read 67426 km. People are stunned when unexpectedly meeting a friend in a remote place, although they are fully unable to quantify the probability of the event.

Recent accounts of the human ability to assess coincidences split into two opposite directions. Some authors would consider that the basic ingredient of coincidences is the probability of the event (Falk, 1989) or of its putative causes (Griffiths & Tenenbaum, 2007), even if they acknowledge the role that descriptive complexity may play (Griffiths & Tenenbaum, 2003). Some other studies have signaled the crucial role of descriptive complexity in cognitive processes (Chater, 1999; Chater & Vitányi, 2003) and more particularly in the perception of coincidences (Feldman, 2004). The former group view "probability as primary, and the relationship between randomness and complexity as a secondary consequence of a statistical inference" (Griffiths & Tenenbaum, 2003). The present account, in line with the second group of studies, puts descriptive complexity at the core of the cognitive competence underlying the perception of coincidences.

In what follows, we first give a list of problems that cognitive models of coincidences must address. We then define the notion of unexpectedness as complexity drop, before showing that this notion nicely accounts for each of the listed problems, including the 'encounter problem'. We finally consider the compatibility of the model with traditional probabilistic accounts.

## What Is To Be Explained

There are required specifications that a cognitive model of coincidence must meet. Let us phrase them using three examples, the double suicide by drowning, the Lincoln-Kennedy parallel and the fortuitous encounter.

- *Analogy*: Each common feature between Lincoln and Kennedy's lives adds to the coincidence. In the case of the two drowned women, interest would have been even higher if they had had the same first name. If they had been of different ages, interest would have been lower. The repeated 100-year interval in the two-president coincidence is an analogy within the analogy.
- *Association*: Non-analogue features, such as the fact that Kennedy was shot in a car named Lincoln, still add to the coincidence value.
- *Prominence*: Kennedy and Lincoln are among the foremost US presidents. The coincidence would be less interesting if it involved obscure US presidents, or prominent Uruguayan presidents (for non-Uruguayan citizens).
- *Round numbers*: The Lincoln-Kennedy coincidence would be slightly less interesting if it involved an 87-year interval.
- *Closeness*: The short distance between the two suicides is a crucial aspect of the coincidence. In the report, proximity in time (early in the same morning), in space (a few kilometers apart) and in social hierarchy (two upper-class women) was highlighted. Conversely, interest in the coincidence would drop if the distance in any of these dimensions was increased.
- *Remoteness*: The remoteness of the place is a crucial ingredient of fortuitous encounters. While there is no coincidence in encountering a colleague two blocks away from workplace, running into a colleague in an obscure village 9000 km away from home is thrilling.
- *Egocentricity*: People are much more sensitive to coincidences (*e.g.* having same birth dates) when they are involved (Falk, 1989). The double suicide is much more intriguing for people who live close to the place where the event took place.
- *Causality*: Surprise vanishes when the coincidental events find a common easy causal explanation, for instance if the two suicidal women knew each other and had made a common decision. Conversely, interest is preserved when the available explanatory hypotheses are highly unlikely (Griffiths & Tenenbaum, 2007) as the fact, mentioned on the radio, that the two drowned women may have influenced each other through telepathy.

All these parameters have *systematic* effects on the perception of coincidences. This is an indication that some genuine cognitive competence is at work and that the surprise generated by coincidences is not the mere side-effect of a general sensitivity to the presence of statistical exceptions. Probabilistic or statistical accounts would fail to account for most of the above properties, though they may suggest local laws in the case of closeness and egocentricity (using Poisson distributions), of remoteness (using a diffusion model) and of explanation (using Bayesian networks: Griffiths & Tenenbaum, 2007). A unified account is proposed now.

## Unexpectedness and Descriptive Complexity

The core property of coincidences is that they are *unexpected*. Unexpectedness has been given various definitions. An obvious one corresponds to Shannon's definition of information $U = \log_2 1/p_i$, where $p_i$ is the *a priori* probability of the event (Shannon, 1948). Another is $U = \Sigma\ p_j^2/p_i$ (Weaver, 1948); it accounts for the fact that improbable events are only interesting if they are a contrast to probable alternatives. Various authors have noticed the close relation between *description complexity*, also known as Kolmogorov complexity, and probability (Solomonoff, 1978; 1997; Chater, 1999; Feldman, 2004). The problem with Solomonoff's definition of probability as $p = 2^{-C}$, where $C$ is the complexity of the event, is that simpler situations are assigned higher probability, as they are more likely to be generated. In coincidences, the reverse happens: Events are perceived as unlikely when they are 'too' simple, for instance simpler than what the 'null hypothesis' predicts (Feldman, 2004). We propose a similar idea, captured in the following definition of *unexpectedness*.

$$U(D) = C_w(D) - C(D) \qquad (1)$$

As observed by Tversky and Koehler (1994), "probability judgments are attached not to events but to descriptions of events". The unexpectedness of an event described through $D$ is the difference between its W-complexity, or complexity 'in the world' (as it is understood by the subject) and its observational complexity or O-complexity. Both $C_w$ and $C$ are cognitive: They correspond to minimal cognitive descriptions available to the individual (Chater, 1999). They express the length of minimal 'programs' given to two different 'machines' that may be dubbed W-machine and O-machine. The W-machine functions according to what the subject sees as the normal behavior of the world (the W-machine, therefore, depends on the subject's knowledge). The O-machine is unconstrained and may use any observational ability available to the subject. $C_w$ and $C$ are the minimal amounts of instructions, in bits, that must be given as input to these machines to *generate* the event described by $D$. These definitions are consistent with the original definition of Kolmogorov complexity, except that the ideal Turing machine is replaced by two versions of the human cognitive analysis capabilities.

Consider the coincidence of seeing 66666 on the clock when driving. The W-machine generates this pattern by copying an uninstantiated digit and then by performing five independent instantiations. The W-complexity thus amounts to $C_w = C_{copy} + 5\times\log_2 10$ (as the complexity of designating one number chosen among $n$ is $\log_2 n$). The O-machine generates the same pattern by copying an instantiated digit: $C = \log_2 10 + C_{copy}$. We thus get: $U = 4\times\log_2 10 = 13.3$ bit.

According to definition (1), merely mentioning 'John F. Kennedy' creates a negative unexpectedness. The world (as the subject knows it) is supposed to be available to the W-machine; therefore, the W-complexity of 'John F. Kennedy' is zero, as he is believed to have existed. The O-machine requires a non-zero input to determine 'John F. Kennedy'. A minimal cognitive description may use a list of famous characters ranked by celebrity. The O-complexity would be the logarithm of Kennedy's rank in that list. Alternatively (especially for people outside the US), a minimal description may go through the mention of the notion of US presidency, and then through the determination of John Kennedy among US presidents. O-complexity here is related to cognitive availability (Tversky & Kahneman, 1973) and may be independently assessed by various means including reaction times.

The mere mention of a feature like Kennedy's age by the time he was killed requires no work from the W-machine, as everyone has an age; but its instantiation to a definite value such as 46 involves some complexity. In the worst case, the W-machine generates that age by copying it from its input. However, the W-machine may determine the age by choosing among likely age values for a US president, which leads to a lower complexity.

## Unexpectedness and Coincidences

Let us show now how the notion of unexpectedness can account for the eight specifications previously listed. We need the notion of *computation sequence*, noted with operator $*$. The sequence $D_1 * D_2$ means that the W-machine needs to generate $D_1$ before generating $D_2$:

$$C_w(D_1 * D_2) = C_w(D_1) + C_w(D_2/D_1) \qquad (2)$$

In a coincidence like the double suicide, the two events $e_1$ and $e_2$ are independent for the W-machine: $C_w(e_2/e_1) = C_w(e_2)$. We get:

$$U(e_1 * e_2) \geq C_w(e_1) + C_w(e_2) - C(e_1) - C(e_2|e_1) \qquad (3)$$

The inequality comes from the fact that the computation sequence may be suboptimal for the O-machine. We can now review the eight specifications.

*Analogy*: This formula correctly predicts the importance of a close analogy between the two events, as best analogies make $C(e_1) + C(e_2|e_1)$ minimum (Cornuéjols, 1996). Each new common element, such as the birth date of the assassins in the Lincoln-Kennedy parallel, is needed twice for the W-machine, but only once for the O-machine, as its contribution to $C(e_2|e_1)$ is zero. Common elements thus add to the unexpectedness of the situation. An additional prediction is that more complex common elements will be more surprising. Interest would indeed grow if the two drowned women wore a tattoo, but even more so if both had a red five-centimeter long tattoo on the right shoulder.

*Association*: Inequation (3) predicts that any element of $e_1$ that the O-machine can reuse to generate $e_2$ will add to the surprise. Hence the mention that Kennedy was killed in car named Lincoln. The W-machine must generate the make of the car and requires several bits which add to $C_w(e_2)$ to distinguish it from other makes; this generation is easier for the O-machine when $e_1$ is given as input, as the name 'Lincoln' is available for free.

*Prominence*: Why is it important that Abraham Lincoln and John F. Kennedy be famous? Formula (3) provides an answer. These two persons exist and are unique in the subject's world; the W-machine has thus no work to do to generate them. For the O-machine, however, their minimal description may go through determining their social role, US president, and then find them in a ranked list of US-presidents. Hence the importance for the coincidence to involve two prominent figures, as their complexity, which amounts to the logarithm of their rank in a list ordered by celebrity, adds to $C(e_1)$ and to $C(e_2|e_1)$. Similarly, a social role such as president of Uruguay, if more complex for the subject, makes the coincidence less surprising, as it adds significantly to $C(e_1)$.

*Round numbers*: The presence of the negative term $-C(e_2|e_1)$ in (3) explains why the Lincoln-Kennedy story is more interesting as it is than if the repeated time interval had been of 87 years instead of 100. The point is that the cognitive complexity of the program that transforms 1846 into 1946 or 1860 into 1960 is simpler than the program that would transform 1846 into 1933 and 1860 into 1947 (the first program affects only one digit). Alternatively, one can observe that 100, as a number, is simpler than 87. The complexity of an integer $n$ is smaller than $\log_2 n$. In the case of round numbers, it is significantly smaller. One million may be concisely defined as $10^6$, or as 1 followed by six copies of 0.

*Closeness*: The role of closeness in time and space, as in the example of the double suicide, can only be understood if the notion of complexity is extended to continuous quantities. Though Kolmogorov complexity is defined only for discrete structures, the complexity of a place is naturally defined as the most concise set of directions that allows finding it. Locating a surface $a^2$ on a two-dimensional area $S$ requires no more than $\log_2 (S/a^2)$ bits. When trying to locate such an area B from a place A, one may use a relative code, in which locations are ranked according to their distance from A. In such a system, the complexity of B amounts to $C(B/A) = \log_2 (\pi d^2/a^2)$, where $d$ is the distance from A to B. This complexity varies as $2 \times \log_2 d$. In expression (3), the W-machine must compute the location of each event independently. The O-machine computes the location of one event, and then may use a relative coding. Hence the crucial importance of the two events being close to each other, as $C(e_2|e_1)$ involves a term $2 \times \log_2 d$. The same prediction holds for the time dimension: Surprisingness decreases as $\log_2 t$ where $t$ is the time interval between the two events.

Proximity effects do not require the presence of two events. On July 22, 2003, a minor blaze on the upper floor of the Eiffel Tower was reported in French national news media. This phenomenon is explained by formula (1). The W-machine must generate the location of the blaze. If fires occur with a spatial density $D$, then localizing one of them demands $\log_2 1/(a^2 D)$ bits, as $1/D$ represents the area of occurrence of one such event on average. The O-machine has first to specify the location of the Eiffel Tower, but this requires minimum complexity as the place is prominent (it is the top most popular monument in France). Then the O-machine gets the location of the blaze for free. The coincidence effect is predicted to decrease as the logarithm of the spatial density of the event and proportionally to the complexity of the monument. No wonder that events happening on prominent places are preferentially reported in the news (Warren, 1934).

*Remoteness*: Fortuitous encounters seem to be an exception to the rule of closeness, as the interest of the coincidence grows this time with the remoteness of the place! A proper application of (3), however, restores the prediction.

First, let us observe that the encounter problem is spectacular, as it provides the best evidence that the human mind is sensitive to description complexity. In this kind of coincidence, interest grows with the *complexity of the place* and with the *simplicity of the encountered person*. The complexity of the place $l$ is the relevant factor, not the distance: a big distant airport may be less complex than the backyard of an obscure building of a lost suburb a few kilometers away. The simplicity of the encountered person $P$ is the relevant factor, not her closeness. Running into a celebrity may be as unexpected as running into a close colleague. These phenomenon is correctly predicted by the fact that unexpectedness varies as $C(l) - C(P)$, as we show now.

Let us compute[1] the unexpectedness of the sequence $ego*P*l(ego)*l(P)$. Here, $l(ego)$ and $l(P)$ designate the presence of self and of $P$ at location $l$. For the W-machine, $C_w(P) = 0$ as $P$ is supposed to exist. In the 'world', $P$'s and $ego$'s presence at $l$ are independent: $C_w(l(P)|l(ego)) = C_w(l(P))$. If $ego$ and $P$ play symmetrical roles, the W-machine requires $C_w(ego*P*l(ego)*l(P)) = 2\,C_w(l(ego))$ to generate the encounter situation. The term $C_w(l(ego))$ corresponds to the minimum size of a set of directions to reach $l$ and amounts to $C(l)$ in most cases (except if $l$ is materially difficult to reach). On the other hand, for the O-machine, $C(l(P)|l(ego))$ is zero. The O-machine demands $C(P)$ to determine $P$ and $C(l(ego))$ to bring $ego$ to $l$. Again, we suppose that $C(l(ego)) = C(l)$. We get, as announced:

$$U(ego*P*l(ego)*l(P)) \geq C(l) - C(P) \quad (4)$$

---

[1] A recursive application of formula (2) gives $C_w(D_1*D_2*D_3) = C_w(D_1) + C_w(D_2/D_1) + C_w(D_3/D_1 \& D_2)$. Irrelevant elements in conditional complexity expressions are omitted for the sake of clarity.

An alternative computation goes over the sequence $ego*l(ego)*l(P)*P$. This time, $P$ is first determined by her position $l(P) = l(ego)$. Now $C_w(l(P)|l(ego))$ is zero, but not $C_w(P/l(P))$. To instantiate $P$, the W-machine has to distinguish among all individuals that may happen to be in $l$. One procedure to do so consists in checking local people first, by delimiting an area of radius $R$ around $l$ and considering all people living within it. Then $R$ is increased until the actual $P$ is reached. In this computation, $l$ and $P$'s home play symmetrical roles, so that $C_w(P/l(P)) = C(l) + c$ (again, $l$ is supposed to be as complex for $P$ as for $ego$). The constant $c$ depends on the spatial density of people. The resulting unexpectedness $C(l) + c - C(P)$ is similar to what we obtained in (4). Note that this second computation still holds when the encounter occurs in the vicinity of $ego$'s home. In this case, $C_w(l(ego))$ and $C(l(ego))$ are negligible, but $C_w(l(P))$ is close to $C(h(P))$ (the complexity of $P$'s home) and the unexpectedness amounts to $C(h(P)) + c - C(P)$. It is indeed quite a coincidence to meet a celebrity in front of one's home.

*Egocentricity*: In the case of fortuitous encounters as for happening to have the same birthday, the coincidence is less impressive if it involves another person $Q$ instead of $ego$ (Falk, 1989). This effect is well predicted by formula (1): The burden of determining $Q$ adds to the O-complexity (but not to the W-complexity) and unexpectedness is diminished by the amount $C(Q)$. This accounts for the fact that first-hand stories are always preferred (Coates, 2003), since they are systematically more unexpected. To show this, we can derive the following inequation from (2):

$$U(D_1*D_2) \geq U(D_1) + U(D_2/D_1) \quad (5)$$

A recursive application of (5) gives: $U(Q*e_1*e_2) = -C(Q) + U(e_1/Q) + U(e_2/Q \& e_1)$. We may consider that $U(e_2/Q \& e_1) = U(e_2/e_1)$, as the most concise description of $e_2$ for the O-machine only uses $e_1$, whereas the W-machine uses neither $e_1$ nor $Q$.

$$U(Q*e_1*e_2) \geq -C(Q) + U(e_1|Q) + U(e_2|e_1) \quad (6)$$

If $Q$ is replaced by $ego$, we get:

$$U(ego*e_1*e_2) \geq U(e_1|ego) + U(e_2|e_1) \quad (7)$$

This explains why coincidences involving $ego$ are more intriguing, as unexpectedness is larger in (7) than in (6). Importantly, this "egocentric touch" (Falk, 1989) is obtained without any extensional reasoning. The present account considers neither alternatives nor "the size of the set to which one implicitly relates" (Falk, 1989).

Equation (7) also explains why a coincidence like the double suicide is more surprising to local people. The value of $U(e_1|ego)$ may be significant, as it would be for a single suicide happening in the vicinity, whereas it may be negative for people living farther away.

*Causality*: Griffith and Tennenbaum (2007) developed a probabilistic (Bayesian) account of the reason why coinci-dences are all the more surprising as they reveal the possibility of an alternative hidden causal structure that is itself highly unlikely. If the less unlikely causal explanation for the double suicide is that the two women influenced each other through telepathy, the feeling of coincidence remains high. If, however, one can believe that they met and decided to commit suicide simultaneously, the coincidence fades away. Our model accounts for the phenomenon, without relying on the notion of probability. Suppose that some causal explanation $H$ accounts for both $e_1$ and $e_2$. This means that $C_w(e_1|H)$ and $C_w(e_2|H)$ are negligible. If $e_1$ and $e_2$ are explained by $H$, then:

$$C_w(e_1*e_2) \leq C_w(H*e_1*e_2) \approx C_w(H) \qquad (8)$$

The W-complexity $C_w(H)$ of a causal explanation $H$ is the amount of information required for the 'world' $W$ to generate $H$. It measures the credibility of the theory (Chaitin, 2004). If $H$ is easy to believe, $U(e_1*e_2)$ does not take significant values and there is no reason to be surprised. Conversely, if $H$ is hard to accept, as for the telepathy hypothesis,[2] then $C_w(H)$ takes high, prohibitive, values. The sequence $H*e_1*e_2$ no longer provides the lowest value for $C_w(e_1*e_2)$ and the hypothesis should be rejected.

## Discussion: What the model predicts

The perception of coincidences is a definite ability, like the ability to decide whether a sentence in one's mother language is grammatical or not. Individuals *know* when a state of affairs makes a good coincidence, and they know if some variation of a given parameter makes a difference. This cognitive ability has to be accounted for, and this is what is attempted here.

The model proposed in this paper meets all the specifications concerning the perception of coincidences that have been listed above, by explaining the systematic role of analogy, of association, of prominence, of round numbers, of closeness, of remoteness, of egocentricity and of causality. Probabilistic models at best predict certain of these features, while the others remain mysterious. The model derived from formula (1) succeeds in predicting them all. Moreover, it does it in a quantitative way. For instance, the effect of spatial closeness is expected to obey a logarithmic law $2 \times \log_2 d$, and time will affect coincidence logarithmically as well. In the case of fortuitous encounters, the complexity (= shortest description) of the place is predicted to be the determining factor, not distance. These various dependencies are all testable and thus make the model falsifiable.

A classical critique addressed to complexity-based approaches is that Kolmogorov complexity is proven to be uncomputable (Li & Vitányi, 1993). Models based on Kolmogorov complexity are thus sometimes believed to lack any real predictive power. This led authors (Simon, 1972; Feldman, 2004) to consider alternative measures of complexity, such as Boolean complexity or algebraic complexity, that are computable. Indubitably, human minds are not ideal computing device that would be free to run any kind of program. Minds are bound to process signals and representations in certain particular ways. However, they are sensitive to the complexity of their own processes, as shown for instance by the existence of strong perceptive biases toward simplicity (Chater, 1999; Chater & Vitányi, 2003). The law encapsulated in formula (1) suggests that this preference for simplicity operates at higher levels of abstraction as well, and across various modalities. Unexpectedness conflates all available sources of complexity (perceptual, structural, spatio-temporal, conceptual) into one single measure that constitutes the judgment of coincidence. This is why the analogy with Kolmogorov complexity is justified, as the only absolute criterion is the size of the shortest *available* description.

One can legitimately worry about the existence of two types of models that address the issue of coincidence perception. If the complexity-based model proposed here is correct, why would probabilistic or Bayesian accounts have partial predictive power? And why are coincidences systematically accompanied by a subjective feeling of low probability (Griffiths & Tenenbaum, 2007)? The relation between descriptive complexity and probability has always been noticed (Solomonoff, 1997), but its usual formulation as $p = 2^{-C(D)}$ is unsatisfactory for our purpose. It would mean that the probability of structures is a decreasing function of their complexity, but coincidences are often striking just because they are simple. To account for the subjective perception of probability in case of lottery results, we introduced an alternative way of deriving probability from complexity (Dessalles, 2006), by defining *cognitive probability* as:

$$p = 2^{-U} \qquad (9)$$

Cognitive probability is defined only for unexpected situations, which means that $U \geq 0$. This definition reverses Shannon's definition, where unexpectedness plays the role of information. It suggests that unexpectedness, as defined by (1), is a more fundamental cognitive notion from which cognitive probability is derived. Formula (9) offers a non-extensional definition of cognitive probability.[3] It explains why individuals do not find that all actual states of the world have zero probability, despite the fact that each is unique. It also explains why certain features are regarded as relevant while others (like the horizontal orientation of dice) are spontaneously discarded. In the case of coincidences, the sensation of low probability appears to result, not from the evaluation of a set of alternatives, but from the perception of a complexity drop. We do not wonder what the alternatives are when we hear that two sixty-year old women chose to walk with their most beautiful dress into the sea early in the same morning (should they have worn pants, or chosen late afternoon?).

---

[2] One way to estimate the W-complexity of the telepathy hypothesis is to imagine the complexity of the conditions that led to an effect in this situation and to the absence of effect in other situations.

[3] Checking whether the probabilities of all elementary alternatives sum to 1 would be quite artificial in most cases, as situations are rarely perceived as exclusive. 66666 km or 67426 km on the clock are respectively read as 'one repeated digit' and 'any amount' and are assigned probability $10^{-4}$ and 1.

## Conclusion

The research initiated in the recent years on the cognitive role of descriptive complexity has already produced valuable results. The model presented in this paper is meant as a contribution to this enterprise. It should be extended in two directions. First, it is important to get a deeper understanding of the processes through which human subjects assess descriptive complexity and to see how they are compatible with the definitions adopted in the present model. Second, it will be interesting to confront the various predictions of the model with actual reactions of subjects to whom various versions of coincidences are presented, as we did for probability estimates (Dessalles 2006). Parameters can be manipulated according to the model to affect surprise. Such studies may prove delicate, however, as surprise cannot be easily elicited successively in the same subjects.

One important conclusion strongly suggested by the model is that the human mind is able to perform non-trivial formal computations. The sensitivity to complexity drop seems to be a general law, which applies across modalities and at all levels of abstraction. This means that our mind is able, in some way, to assess the complexity of its own information processing. Such a possibility opens many new perspectives.

**www.simplicitytheory.org**